\documentclass[letterpaper, 10 pt, conference]{ieeeconf}  

\IEEEoverridecommandlockouts                              

\usepackage{amsfonts}
\usepackage{amsmath, float}
\usepackage{graphicx}

\usepackage{colortbl}
\usepackage{booktabs}
\graphicspath{{Figures/}}
\usepackage{amssymb,bbm, array}
\usepackage[table,xcdraw]{xcolor}
\usepackage{multirow}
\usepackage{adjustbox}
\usepackage{multirow}
\usepackage{bbding, pifont}
\usepackage[hang,flushmargin]{footmisc}
\usepackage[breaklinks,colorlinks]{hyperref}

\definecolor{darkred}{rgb}{0.7, 0, 0}
\definecolor{darkblue}{rgb}{0.0, 0.0, 0.7}
\definecolor{darkgreen}{rgb}{0, 0.4, 0}

\newcommand{\cmark}{\textcolor{darkgreen}{\ding{51}}}
\newcommand{\xmark}{\textcolor{darkred}{\ding{55}}}

\usepackage{cite}
\let\labelindent\relax
\usepackage{enumitem}
\usepackage{tabularray}
\usepackage[hang,flushmargin]{footmisc}
\usepackage[breaklinks,colorlinks]{hyperref}

\UseTblrLibrary{siunitx}
\usepackage[misc]{ifsym}

\title{\LARGE \bf
RoboSwap: A GAN-driven Video Diffusion Framework For Unsupervised \textit{Robo}t Arm \textit{Swap}ping
}

\author{
  Yang Bai$^{1,2}$, Liudi Yang$^{2,4}$, George Eskandar$^{2}$, Fengyi Shen$^{2,3}$, Dong Chen$^{2}$, \\
  Mohammad Altillawi$^{2}$, Ziyuan Liu$^{2\textrm{ \Letter}}$\thanks{\textrm{\Letter} : Corresponding author.}, Gitta Kutyniok$^{1}$ \\
  $^{1}$ Ludwig-Maximilians-Universität München (LMU Munich), 
  $^{2}$ Huawei Heisenberg Research Center, \\
  $^{3}$ Technische Universität München (TUM), 
  $^{4}$ Albert-Ludwigs-Universität Freiburg
}

\begin{document}

\maketitle
\thispagestyle{empty}
\pagestyle{empty}


\begin{abstract}
Recent advancements in generative models have revolutionized video synthesis and editing. However, the scarcity of diverse, high-quality datasets continues to hinder video-conditioned robotic learning, limiting cross-platform generalization. In this work, we address the challenge of swapping a robotic arm in one video with another— a key step for cross-embodiment learning. Unlike previous methods that depend on paired video demonstrations in the same environmental settings, our proposed framework, RoboSwap, operates on unpaired data from diverse environments, alleviating the data collection needs. RoboSwap introduces a novel video editing pipeline integrating both GANs and diffusion models, combining their isolated advantages. Specifically, we segment robotic arms from their backgrounds and train an unpaired GAN model to translate one robotic arm to another. The translated arm is blended with the original video background and refined with a diffusion model to enhance coherence, motion realism and object interaction. The GAN and diffusion stages are trained independently. Our experiments demonstrate that RoboSwap outperforms state-of-the-art video and image editing models on three benchmarks in terms of both structural coherence and motion consistency, thereby offering a robust solution for generating reliable, cross-embodiment data in robotic learning.


\end{abstract}

\section{INTRODUCTION}
The increasing interest in embodied intelligence has catalyzed extensive research into the use of video-based learning methodologies for training robotic arms \cite{learning2imitate,learningbywatching,Learningmulti-stage,sharma19thirdperson,schmeckpeper2020rlv,skill}. Recent frameworks such as Vid2Robot\cite{vid2robot} underscore the potential of video-conditioned learning, enabling robotic systems to develop adaptive policies by observing a wide range of actions in real-world settings. However, existing online datasets are constrained in terms of scope and operational context, limiting the ability to train robotic arms across diverse scenarios. Generative models hold the promise to augment the size and diversity of robotic datasets, reducing manual data collection and enabling large-scale robotic learning across different tasks, environments and embodiments.

\begin{figure}[ht!]
  \centering
  \includegraphics[width=0.48\textwidth]{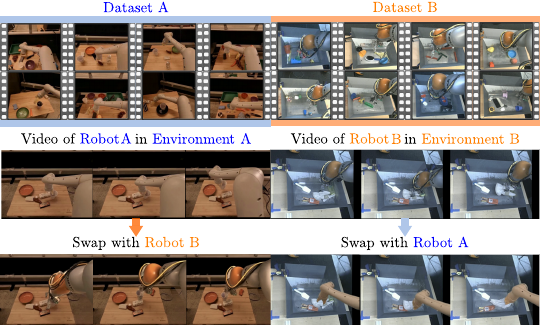}
  \caption{We propose RoboSwap, a GAN-driven Video Diffusion Approach to swap the robot arm in a video with another one. Contrary to previous works~\cite{humantrans, mirage} which swap the robot arms in the same environment, our approach enables the replacement of robotic arm across different domains, without requiring paired datasets. The generated videos conform to the motion dynamics of the reference robotic arm.}
  \label{fig:teaser}
  \vspace{-2em}
\end{figure}
In this paper, we address the problem of swapping a robotic arm A with another robot B in the videos with a focus on preserving the motion, to enable robot arm B to perform the same task as A (Fig.~\ref{fig:teaser}). This approach is particularly valuable in scenarios where task videos from robotic arms of different manufacturers are available, but the goal is to deploy a custom robotic arm in a new environment. Reconstructing the original environment and manually collecting the data (e.g., through teleoperation) is impractical, but replacing the robot in the original video enables to transfer the learned behavior across both embodiments and environments by learning from the newly generated video.

Swapping a robot in a video with another robot presents several significant challenges, particularly when addressing the task in real-world scenarios, as opposed to controlled simulation environments. One major difficulty is the lack of paired data— synchronized video demonstrations of both robots performing the same task in identical environments are costly and time-consuming to obtain. In practice, however, only videos of robots A and B performing tasks in different environments are available, which makes supervised training unfeasible due to the lack of ground truth. Furthermore, for the swapped robot to maintain realistic motion, it must not only follow the trajectory of robot A but also adhere to physical constraints, such as proper object grasping. This is particularly challenging when the two robots differ structurally, with varying joint configurations that complicate the transfer of motion. Additionally, the videos of robots A and B are often captured in different environments with distinct objects, giving rise to domain gaps, such as the sim-to-real gap and environmental location discrepancies. These challenges must be overcome to enable realistic and accurate robot swapping in diverse, real-world scenarios.

Recently, diffusion models~\cite{ddpm} have emerged as powerful generative tools, demonstrating exceptional performance in producing high-quality, diverse samples. However, the problem of robotic arm swapping in videos remains unresolved. Some frameworks~\cite{i2vedit,ku2024anyv2v} allow local edits within a video, but they are typically trained on general-purpose datasets with limited interaction between a single subject and the environment. In contrast, embodied AI data involve multiple subjects interacting in complex environments. Other frameworks can achieve robot-to-robot transfer but only with paired data~\cite{mirage,humantrans}. This makes us raise the following question: \textit{How to generate a video with a swapped robotic arm without having to collect paired data for each new robot?} Answering this question can help alleviate the training requirements for transfer learning across different embodiments and environments.

Generative Adversarial Networks (GANs) are an older line of work that have particularly excelled in unpaired translation between images or videos~\cite{cyclegan,cutgan} as they do not have the gaussian noise assumption like diffusion models. Unpaired GANs can learn to reasonably translate from one domain to another even when the shape of the translated object changes between the two datasets. However, they lag behind diffusion models in sample quality, diversity, and inpainting ability. 

\textbf{Contribution}. In this work, we propose a novel framework, RoboSwap, which can swap Robot A in a video with another Robot B while preserving the original motion and environment. Our key insight is to combine the benefits of GANs and diffusion models in a novel framework bringing the best of each approach: the unsupervised translation ability of GANs, and the photorealism of video diffusion models. We break down the problem into two tasks: robot-to-robot translation, and video inpainting with the new robot. In the first stage, we segment the robotic arms in the two datasets and learn an unpaired GAN to translate from one robot to another, preserving the alignment and pose of the robot during translation. This particularly compensates for the lack of paired cross-embodiment data. In the second stage, we propose a diffusion-based approach to insert a robot in a specific environment. Both stages are trained separately, then, combined during inference in an end-to-end manner. In summary, our contributions are as follows.
\vspace{-1em}

\begin{enumerate}
    \item We introduce RoboSwap, a method for swapping a robotic arm in a video with another one in an unpaired way, alleviating the data collection requirements for cross-embodiment transfer learning.
    \item We propose a novel combination of GANs and diffusion models in a two-stage approach, for which we ablate several design choices, demonstrating their impact on the model’s performance.
    \item We achieve state-of-the-art results compared to both diffusion models and GANs across three different benchmarks, showcasing the generalizability of our approach.
\end{enumerate}

\begin{figure*}[ht]
  \centering
  \vspace{1em}
  
  \includegraphics[width=\textwidth]{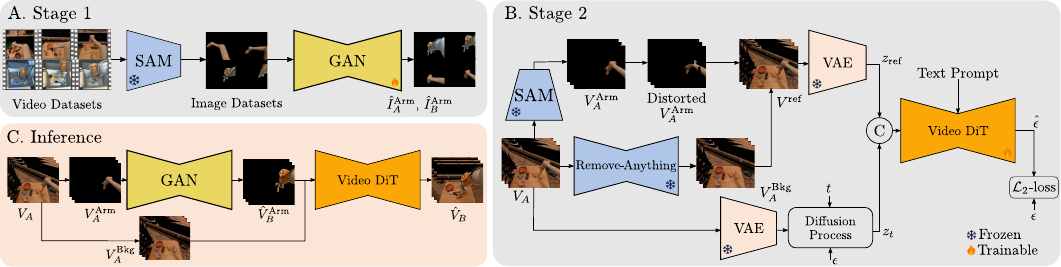}
  \caption{RoboSwap has two stages, each trained independently. (A) First, we train a GAN~\cite{cyclegan} on two unpaired datasets of robotic arms, segmented from their backgrounds, to eliminate background interference and focus on pose translation between the robots only. (B) Second, A video diffusion model is then trained to refine the  blended inputs of background and translated robotic arms. To mitigate GAN artifacts, distortion augmentations are applied to the arms during training. The diffusion model is trained on refining the videos in a self-supervised way on both datasets A and B. (C) During inference, the robotic arm is segmented, transformed via GAN, composited onto the background, and refined by the video diffusion model for enhanced realism. }
  \label{fig:method}
  \vspace{-1.5em}
\end{figure*}

\section{RELATED WORK}

\subsection{Unsupervised Image and Video Generation with GANs}

GANs have shown remarkable success in unsupervised learning, particularly in tasks where paired training data is unavailable. Unlike supervised approaches that rely on labeled data, GANs methods such as CycleGAN\cite{cyclegan} CUTGAN\cite{cutgan} utilize an adversarial training framework to learn data distributions without explicit supervision. Regarding video generation and editing, GAN-based methods extend these capabilities to temporal domains~\cite{videogan, mocogan, tgan}. Despite these advancements, existing video GAN models often struggle with long-term consistency, complex motion representation, and structural integrity, particularly in tasks requiring precise alignment with reference motion. In our work, we use unpaired GAN models to learn robot-to-robot translation, but segment the robotic arms from their background to eliminate unwanted interference and focus on pose translation.


\subsection{Diffusion-based Models for Image and Video Editing}

The rapid development of diffusion models \cite{zhang2023adding,stablediffusion,ho2020denoising} has significantly enhanced image and video editing capabilities. Text-to-image diffusion models provide a convenient way for users to modify images through natural language descriptions. However, these methods still face limitations in precise control, such as difficulties in fine-grained editing of specific objects or seamless integration into complex scenes. Image-based editing methods \cite{gal2022textual, ruiz2022dreambooth} aimed at addressing these issues by taking reference images and textual prompts as input for image generation. However, they typically require extensive fine-tuning and often struggle to integrate objects naturally into new scenes. Such challenges are particularly pronounced in complex tasks like robotic arm replacement, where issues such as hand/gripper distortion or object shape anomalies lead to unrealistic results. Although some excellent work has been done in object replacement~\cite{hoiswap}, the task of replacing robotic arms remains an unmet need.

Diffusion-based methods have been extended to the time domain making video editing feasible \cite{ramesh2021dalle,hong2022cogvideo,xing2023makeyourvideo,khachatryan2023text2videozero}. However, existing methods often rely on time-consuming video decomposition or costly per-video fine-tuning, limiting their practicality. Consequently, training-based video editing methods have become a research focus. These approaches leverage large-scale datasets for pre-training, enabling efficient editing during inference. Moreover, existing video editing techniques commonly rely on structural signals such as depth maps and optical flow to control motion~\cite{revideo}, which preserve the motion patterns of the original video object and fail to effectively adapt to the motion control requirements of reference objects.

To overcome the limitations of existing approaches in shape transformation and motion alignment, as well as the lack of paired video datasets for training, we propose a GAN-driven video diffusion model, which utilizes cross-domain arm videos generated from GANs to build input to prompt the video diffusion model, thus performing unsupervised robot arm swapping in videos.

\subsection{Robot Swapping in Videos}

Existing methods for robot swapping either focus on robot-to-robot transfer within the same environment \cite{mirage, humantrans} or cross-embodiment imitation within fixed tasks \cite{cross_grasp}, but struggle when both robotic embodiment and task domain vary simultaneously. For example, the Mirage framework \cite{mirage} uses a cross-painting technique to transfer simulated robotic arms within the same environment, but it is limited to the simulator domain and does not extend to real-world applications. Similarly, Bharadhwaj et al. proposed Translating Human Interaction Plans \cite{humantrans}, where human hands are inpainted into robotic task environments, enabling transfer from human demonstrations to robotic execution. Their approach requires collecting paired human-robot demonstrations, limiting its generalizability. Instead, our approach only requires two unpaired video datasets. 
\vspace{-0.5em}

\section{METHOD}

We propose \textit{RoboSwap}, an unsupervised framework comprising two training stages (a GAN stage and a diffusion stage), as well as a hybrid GAN-driven diffusion inference stage, with the aim of transferring robotic arm appearance, structure (geometry), and motion across different robotic arm domains and different environments (Fig.~\ref{fig:method}).

Specifically, in the first stage, we take disentangled robotic arms and leverage unsupervised GAN training on the unpaired video frames to learn  a mapping across different robotic arms while preserving their poses. In the second stage, we train a video diffusion model to inpaint a given robotic arm into the video stream of a given environment. During inference, combining GAN-based arm transfer and diffusion-based arm-to-environment inpainting, we achieve high-quality unpaired robotic arm swapping in videos. Our unsupervised \textbf{\textit{RoboSwap}} framework even succeeds in the most challenging scenario, where not only the robotic arms but also their background environments are different.
Below, we illustrate each stage of our framework.

\subsection{Stage 1: Unsupervised Robot-to-Robot Translation}
Existing image or video editing methods are ill-suited to the cross robotic arm transfer tasks due to the absence of paired training data, i.e., the target robotic arm domain does not contain a corresponding ground truth for an input source arm. To circumvent this limitation,  we propose to take advantage of the generative capabilities and unsupervised learning properties of GANs to map one source robotic arm into the desired target arm domain. To train a GAN for this aim, we curate domain separated datasets of robotic arm images from the provided videos and disentangle them out from their  background environments by Grounded-SAM \cite{ren2024grounded} and TrackAnything\cite{yang2023track}. For given video sequences \(V_A=\{I_{Ai}\}_{i=1}^N\) and \(V_B=\{I_{Bi}\}_{i=1}^N\), each consisting of $N$ camera frames $I$, depicting robotic environments A and B, respectively, we obtain a set of corresponding binary masks $\{M_{Ai}\}_{i=1}^N$ and $\{M_{Bi}\}_{i=1}^N$, corresponding to the camera frames of the video sequences, where $1$ denotes the robotic arm and $0$ otherwise. We use these masks to extract the robotic arms from the environment background forming robotic arm videos \(V^{\textrm{Arm}}_A=\{I^{\textrm{Arm}}_{Ai}\}_{i=1}^N\) and \(V^{\textrm{Arm}}_B=\{I^{\textrm{Arm}}_{Bi}\}_{i=1}^N\) as shown in Fig.~\ref{fig:method}.

To learn this unsupervised robotic arm transfer from the source domain (i.e., \(I^{\textrm{Arm}}_{A}\)) to the target domain (i.e., \(I^{\textrm{Arm}}_{B}\)), we use CycleGAN \cite{cyclegan} as it is fast and easy to train and allows real-time inference. During training, random pairs of images depicting the different robotic arms were selected from the formed dataset to generate the translated images \(\hat{I}^{\textrm{Arm}}_{A}\) and \(\hat{I}^{\textrm{Arm}}_{B}\).

The loss of the GAN training stage is defined as:
\begin{align*}
L_{\text{GAN}} &= L^{A\rightarrow B}_{\text{adv}}(I^{\textrm{Arm}}_{A}, \hat{I}^{\textrm{Arm}}_{B}, I^{\textrm{Arm}}_{B}) \\
&\quad + L^{B\rightarrow A}_{\text{adv}}(I^{\textrm{Arm}}_{B}, \hat{I}^{\textrm{Arm}}_{A}, I^{\textrm{Arm}}_{A}) \\
&\quad + \lambda L_{\text{cyc}}(I^{\textrm{Arm}}_{B}, \tilde{I}^{\textrm{Arm}}_{B},  \tilde{I}^{\textrm{Arm}}_{A}, I^{\textrm{Arm}}_{A}),
\end{align*}
where \(L^{A\rightarrow B}_{\text{adv}}\), \(L^{B\rightarrow A}_{\text{adv}}\), and \(\lambda\) are the adversarial losses across robotic arm domains, \(L_{\text{cyc}}\) are the cyclic reconstruction losses and \(\lambda\) is their weighting factor, and \(\tilde{I}^{\textrm{Arm}}_{B}\), \(\tilde{I}^{\textrm{Arm}}_{B}\) are the cyclic reconstructed arms. Detailed definitions of the losses are presented in \cite{cyclegan}.

During inference, our GAN outputs provide an important inter-domain mapping as input to the diffusion model to compensate for the lack of paired data across different robotic arm domains.

\subsection{Stage 2: Robotic Arm Inpainting}

The next stage is to train a model which inpaints a given robotic arm into a certain environment (video), while following the original robot motion as indicated by the prompt. We propose a video diffusion approach to achieve that goal as we notice that pre-trained video diffusion models like CogVideo~\cite{yang2024cogvideox} can serve as a robust backbone for robotic arm inpainting, owing to their versatile pretrained knowledge (see Fig.~\ref{fig:method}-B).

\noindent\textbf{Constructing Data for Video Diffusion Training:}
We propose to setup the input data as decoupled foreground and background streams and train the video diffusion to blend them both in one coherent video stream. For the foreground data preparation, we follow a similar approach as in Stage 1 to extract robotic arm videos. Specifically, we compute binary masks for the video frames \(V_A\) and extract a video of the robotic arm from the given environments \(V_A^{\textrm{Arm}}\), \(V_B^{\textrm{Arm}}\). 

To perform unsupervised robotic arm swapping, our aim is to feed the video diffusion model with GAN transferred videos during the inference stage. 
However, unsupervised GAN models are known to suffer from degraded output quality with adversarial distortions. Thus, there exists a distribution gap between training time and test time. At training time, arm videos directly segmented from real video datasets are used while at test time, the GAN output is used. Therefore, we introduce a further step to pre-process the reference arm videos \(V^{\textrm{Arm}}\). To mimic similar GAN artifacts on the reference arm videos during training, we apply small data distortions including elastic transformations, perspective transformations, and Gaussian blurring to obtain a distorted robot video sequence \(V_{\textrm{Arm}}\). Simultaneously, we process the source video sequence \(V_A\) using Remove-Anything \cite{yu2023inpaint} to extract its background environment \(V^{\textrm{Bkg}}_A\). 
Thus the data used to train this stage consists of triplets ($V_A^{\textrm{Arm}}$,  $V_A^{\textrm{Bkg}}$, $V_A$) and ($V_B^{\textrm{Arm}}$,  $V_B^{\textrm{Bkg}}$, $V_B$).  

\noindent\textbf{Video Inpainting Training:} We design the model as a video-conditioned Latent Diffusion Model (LDM), that performs the mapping $V_A^{\textrm{Arm}} + V_A^{\textrm{Bkg}} \rightarrow V_A$ . We alpha blend both streams, the distorted foreground and the background videos into one video, denoted by $V_{\textrm{ref}}$, that is passed to the diffusion model to be refined. Note that $V_{\textrm{ref}}$ suffers from clear misalignments during training (an intentional design choice by adding distortions on $V_A^{\textrm{Arm}}$) and during inference (because of the imperfect GAN output), such as unrealistic motion and unrealistic object interactions.  


To encode the videos into a latent space, we employ a pre-trained open-source 3D Causal VAE encoder $\mathcal{E}$. For a video $V$ of dimension $\mathbb{R}^{3 \times N \times H \times W}$, where $N$ represents the number of frames, and $H$ and $W$ denote the height and width of the frames, respectively the VAE processes the video to obtain latent feature $Z$ of dimension $\mathbb{R}^{16 \times N \times H_{l} \times W_{l}}$, where $H_l$ and $W_l$ refer to the height and width in the latent space. To reconstruct the source video (ex \(V_A\)), we first process it by the VAE to obtain latent feature $z_A$ and apply the forward diffusion process by progressively adding Gaussian noise to $z_A$ over $T$ steps, generating a sequence of noisy samples $\{z_{A0}, z_{A1}, \dots, z_{AT}\}$.

We further incorporate two conditional signals in addition to the standard noisy latent variable $z_t$ and the diffusion time step $t$. We process the reference video \(V^{\textrm{ref}}\) using the VAE to produce a latent feature $Z^{\text{\textrm{ref}}} \in \mathbb{R}^{16 \times N \times H_l \times W_l}$. This latent feature $Z^{\text{\textrm{ref}}}$ is then concatenated channel-wise with the original latent variable $z_A$ before being fed into the denoising network $\epsilon_{\theta}$. In addition, we include a robotic arm task description in the form of a textual prompt. We encode the text prompt using the T5 model~\cite{t5}, yielding a vector $d_c \in \mathbb{R}^{4096}$. This vector provides high-level contextual guidance and is incorporated into the denoising network $\epsilon_{\theta}$ via cross-attention.

We thus train the denoising network $\epsilon_{\theta}(\cdot, t)$, implemented as a DiT, to learn to progressively remove the added noise and reconstruct the video sequence. In practice we do this by adding some LoRA parameters \cite{hu2022lora} to finetune CogVideoX base model. The training objective of Stage 2 is:
\[
L_{\text{diffusion}} = \mathbb{E}_{z, z^{\textrm{ref}}, d^c, \epsilon \sim \mathcal{N}(0,1), t} \left\| \epsilon - \epsilon_{\theta}(z_t, z^{\textrm{ref}}, d^c, t) \right\|_2^2
\]

\subsection{Hybrid Inference Stage: Robotic Arm Swapping}
We build our hybrid inference pipeline for robotic arm swapping as follows (see Fig.~\ref{fig:method}-C). To swap robotic arm video \(V_{B}\) from A into B with background environment video \(V^{\textrm{Bkg}}_A\), we first obtain the GAN swapped arm video \(\hat{V}^{\textrm{Arm}}_{B}\) from \(V^{\textrm{Arm}}_{A}\), then feed it to the video diffusion model together with \(V^{\textrm{Bkg}}_A\) to generate \(\hat{V}_{B}\) which can execute the same task. This denoising diffusion process can be described as,

\begin{align*}
\hat{V}^{B} = \int_{0}^{T} \mathcal{N}(\mathbf{V}_{t-1}; \mu_\theta(\mathbf{V}_t, V^{\textrm{Bkg}}_A, \hat{V}^{\textrm{Arm}}_{B}, t), \sigma_t^2 \mathbf{I}) \, dt.
\end{align*}
where \( \mathbf{V}_t \) represents the noisy video state at timestep \( t \), transitioning from \( t = T \) to \( t = 0 \), \( \mu_\theta(\mathbf{V}_t, V^{\textrm{Bkg}}_A, \hat{V}^{\textrm{Arm}}_{B}, t) \) is the mean predicted by the model at each timestep, estimating the denoised video \( \mathbf{V}_{t-1} \) while incorporating information from both conditioning inputs. The variance \( \sigma_t^2 \) determines the noise level at each step. \( \mathcal{N}(\mathbf{V}_{t-1}; \mu_\theta, \sigma_t^2 \mathbf{I}) \) defines a Gaussian distribution centered at \( \mu_\theta(\mathbf{V}_t, V^{\textrm{Bkg}}_A, \hat{V}^{\textrm{Arm}}_{B}, t) \) with covariance \( \sigma_t^2 \mathbf{I} \), representing the probability distribution of the denoised video at the previous timestep. The integral accumulates these updates over time, with \( dt \) denoting an infinitesimal timestep in the continuous denoising process. Note that the diffusion sampling is still operated in the latent space, which is omitted here for simplicity.

\section{EXPERIMENTS AND RESULTS}

To comprehensively evaluate the results of robotic arm swapping, we consider both image editing (applied only to the first frame of the video) and video editing. In Section 4.1, we describe the experimental setup, and in Section 4.2, we present the results and comparative analysis.
\subsection{Experimental setup}

\noindent\textbf{Datasets}
We utilize three robotic arm datasets, two of which are publicly available: BCZ (Google Robot)\cite{bcz}, QT-Opt (Kuka Robot)\cite{kuka}. The third dataset was collected privately using a UR5 robotic arm. We establish three benchmarks: Google $\rightarrow$ Kuka, Kuka $\rightarrow$ Google, and Google $\rightarrow$ UR5. In the first stage, 10,000 images of each dataset (i.e. Google Robot and Kuka, or Google Robot and UR5) are used to train the GAN. In the second stage, the model is trained with around 32.6K videos, collected from all datasets. The evaluation set for each benchmark comprises 100 videos, each with a corresponding robotic arm replacement.

\noindent\textbf{Baseline Models}  
In image editing, our first-stage model is compared with two powerful inpainting models, PBE\cite{yang2023paint} and AnyDoor\cite{chen2024anydoor}. For video editing, we consider the following baselines:

\begin{enumerate}
    \item Two image editing baselines (\textbf{PBE} \cite{yang2023paint} and \textbf{AnyDoor} \cite{chen2024anydoor}) to replace every frame from the original video.
    \item Applying the best image editing results (including our approach) for \textbf{AnyV2V}\cite{ku2024anyv2v}, a tuning-free framework that takes one edited frame as conditional guidance.
    \item Same input in (2) applied into \textbf{I2VEdit}\cite{i2vedit}, which propagates edits from a single frame with Coarse Motion guidance to the entire video.
    
\end{enumerate}

We also add these additional baselines, all based on variations of our models. Many of these baselines use the same video diffusion model (CogVideoX) for fair comparison.

\begin{enumerate}
    \item We use \textbf{CycleGAN} model trained on the whole videos (foreground + background).
    \item \textbf{I2V-Original} We retain the original robotic arm in the first frame and generate the video with CogVideoX-Image-to-Video (I2V) model by changing the robotic arm type in the prompt.
    \item \textbf{I2V-Bkg} We use RemoveAnything to remove the robotic arm from the first frame, then generate the video with CogVideoX-I2V by changing the robotic arm type in the prompt.
    \item \textbf{I2V-Swapped} Best image editing method applied to the first frame and we use CogVideoX-I2V to generate the video. We also change the robotic arm type in the prompt.
\end{enumerate}

\noindent\textbf{Evaluation:}  
We evaluate the models using both human and automatic metrics. The user study involved 15 participants, who assessed 100 image edits and 25 video edits.\\
\textbf{1. Automatic Evaluation:} We use the following quantitative metrics for automatic evaluation: Motion Smoothness(\textbf{mot. Smth.}), Background consistency (\textbf{bg. Cons.}), Subject Consistency(\textbf{subj. Cons}), Temporal flickering (\textbf{tem. Fli.}). The metrics are calculated using VBench\cite{vbench} to assess the overall quality of the videos.\\
\textbf{2. User Study:} For image editing, participants are required to judge whether the robotic arm has been swapped in a realistic way. For video editing, participants must answer this question and additionally select 3 preferred results. In the survey, we display 10 editing results, with 8 from the baseline methods and 2 versions of our model (by using different GAN models in Stage 1, namely CycleGAN~\cite{cyclegan} and CUT~\cite{cutgan}). All user preference results are presented in percentage format. Participants are asked to choose the preferred results based on robotic arm replacement success, interaction realism with objects, motion alignment and overall quality (for video editing). Samples are randomly shuffled during the survey.\\


\begin{table}[!ht]

\begin{center}
\setlength{\tabcolsep}{3.0pt}
\begin{tabular}{ c c c c}
\hline
Method & Google $\rightarrow$ Kuka & Kuka $\rightarrow$ Google & Google $\rightarrow$ UR5\\
Swapped & Swapped &  Swapped & Swapped\\
\hline
PBE & \xmark & \xmark & \xmark \\ 
AnyDoor & \xmark & \xmark & \xmark \\ 
\hline
\textit{Ours} & \cmark & \cmark & \cmark \\ 

\end{tabular}
\caption{Comparison against different Image editing Approaches using user study: \textit{if it can change the robot correctly?}}
\label{tab:image_comparison}
\end{center}
\vspace{-4em}
\end{table}

\begin{figure}[ht!]
  \centering
  \includegraphics[width=0.48\textwidth]{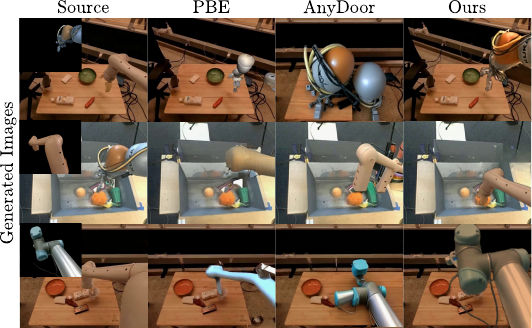}
  \caption{COMPARISON AGAINST DIFFERENT IMAGE EDITING APPROACHES.}
  \label{fig:imgedit_comparison}
\end{figure}

\noindent\textbf{Implementation details:} Stage 1 model is trained for 500 epochs on 256×256 resolution images. For stage 2 training, the input video resolution is set to 480×720, with a latent space of 60×90, where 49 frames are output at 8 frames per second (fps), and the model is trained for 100K steps. We train each stage on 8 NVIDIA-A100-80G GPU about 3 days.

\subsection{Editing Results}

\noindent\textbf{Quantitative evaluation}
RoboSwap outperforms the other baselines across all automatic metrics and in the human evaluation (see Upper Part of Table \ref{tab:main_comparison}).


\begin{table*}[!ht]
    \centering
    \vspace{1em}
    
    \setlength{\tabcolsep}{3pt}
    \renewcommand\arraystretch{1}
    \adjustbox{width=\linewidth}{\begin{tabular}{l| cccccc | cccccc | cccccc }
    \hline
        \multirow{2}{*}{Method} & \multicolumn{6}{c|}{Google robot$\rightarrow$Kuka robot} & \multicolumn{6}{c|}{Kuka robot$\rightarrow$Google robot} & \multicolumn{6}{c}{Google robot$\rightarrow$UR5} \\
        ~ & Mot. & Bg.  & Subj. & Tem. & Can be &  User & Mot. & Bg. & Subj.  & Tem. & Can be &  User & Mot. & Bg. & Subj. & Tem. & Can be &  User \\
        ~ & Cons. & Cons. & Cons. & Fli. & Swapped &  Pref. & Cons. & Cons. & Cons. & Fli. & Swapped & Pref. & Cons. & Cons. & Cons. & Fli. & Swapped &  Pref. \\
        \hline
        PBE (Per-frame) & 0.9777 & 0.9119 & 0.8455 & 0.9761 & \xmark &  0 & 0.9729 & 0.9336 & 0.8739 & 0.9694 & \xmark &  0 & 0.9787 & 0.9345 & 0.8806 & 0.9758 & \xmark &  0 \\
        AnyDoor (Per-frame) & 0.9423 & 0.8956 & 0.8605 & 0.9343 & \xmark &  0 & 0.9381 & 0.8826 & 0.8458 & 0.9314 & \xmark &  0 & 0.9185 & 0.9007 & 0.8639 & 0.9093 & \xmark &  0 \\
        AnyV2V & 0.9869 & 0.9213 & 0.8732 & 0.9830 & \cmark &  13.99 & 0.9832 & 0.9000 & 0.8845 & 0.9806 & \cmark &  3.28 & 0.9872 & 0.9129 & 0.9174 & 0.9825 & \cmark &  6.74 \\
        I2VEdit & 0.9864 & 0.9479 & 0.9139 & 0.9726 & \cmark &  20.69 & 0.9881 & 0.9465 & 0.9374 & 0.9866 & \cmark &  4.56 & 0.9839 & 0.9459 & 0.9104 & 0.9755 & \cmark &  17.52 \\
        \midrule
        CycleGAN & 0.9822 & 0.9492 & 0.9247 & 0.9765 & \xmark & 0 & 0.9767 & 0.9473 & 0.9274 & 0.9730 & \xmark & 0 & 0.9747 & 0.9416 & 0.9400 & 0.9656 & \xmark & 0 \\
        I2V-Original & 0.9895 & 0.9414 & 0.9230 & 0.9884 & \xmark & 0 & 0.9904 & 0.9482 & 0.9312 & 0.9924 & \xmark & 0 & 0.9908 & 0.9612 & 0.9484 & 0.9906 & \xmark & 0\\
        I2V-Bkg & 0.9938 & 0.9535 & 0.9416 & 0.9946 & \xmark & 0 & 0.9922 & 0.9537 & 0.9535 & 0.9945 & \xmark & 0 & 0.9936 & 0.9527 & 0.9412 & 0.9946 & \xmark & 0\\
        
        I2V-Swapped & 0.9875 & 0.9268 & 0.9413 & 0.9857 & \xmark  & 8.64& \textbf{0.9944} & \textbf{0.9824} & 0.9473 & \textbf{0.9960} & \cmark  & 30.47& 0.9869 & 0.9455 & 0.9116 & 0.9857 & \xmark & 19.24\\

        \midrule
        \textit{Ours with CUT} & 0.9889 & \textbf{0.9645} & \textbf{0.9413} & \textbf{0.9892} & \cmark & 27.97 & 0.9927 & 0.9638 & 0.9560 & 0.9946 & \cmark & 29.13 & 0.9905 & 0.9588 & 0.9549 & \textbf{0.9926} & \cmark & 24.62\\
        
        \textit{Ours with CycleGAN} & \textbf{0.9894} & 0.9617 & 0.9381 & \textbf{0.9892} & \cmark  & \textbf{28.71} & 0.9934 & 0.9558 & \textbf{0.9573} & 0.9952 & \cmark & \textbf{32.56} & \textbf{0.9906} & \textbf{0.9628} &\textbf{ 0.9737} & \textbf{0.9926} & \cmark & \textbf{31.88}\\

    \end{tabular}}
    \caption{Comparison of Different Video Editing Approaches.Bold refers to the best results among methods that  swapped the robot correctly. Upper Part:Baselines, Lower Part: Ablations.}
    \label{tab:main_comparison}

\end{table*}

\begin{figure*}[ht!]
  \centering
  \includegraphics[width=1\textwidth]{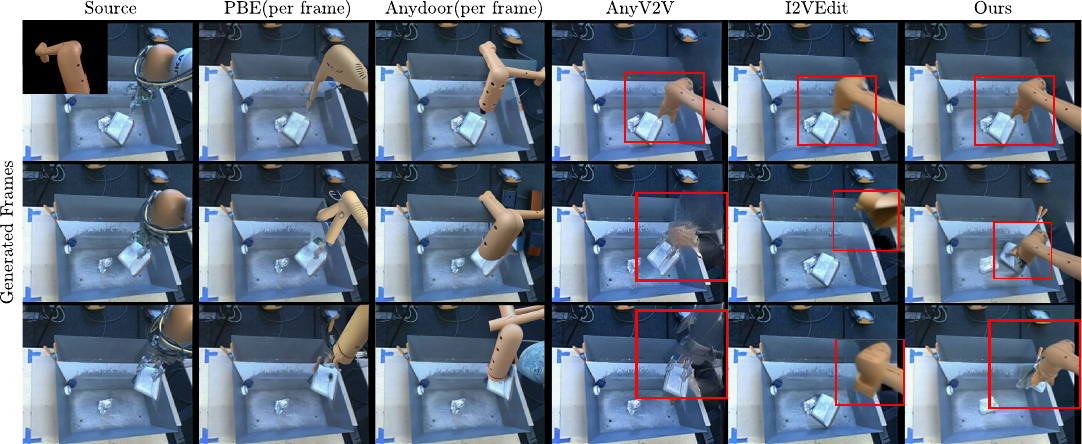}
  \caption{Comparison against different Video Editing Approaches for Robot Swap on the Kuka Robot → Google Robot task.}
  \label{fig:main_comp_qual}
\vspace{1em}
\end{figure*}

\begin{figure*}[ht!]
  \centering
  \includegraphics[width=1\textwidth]{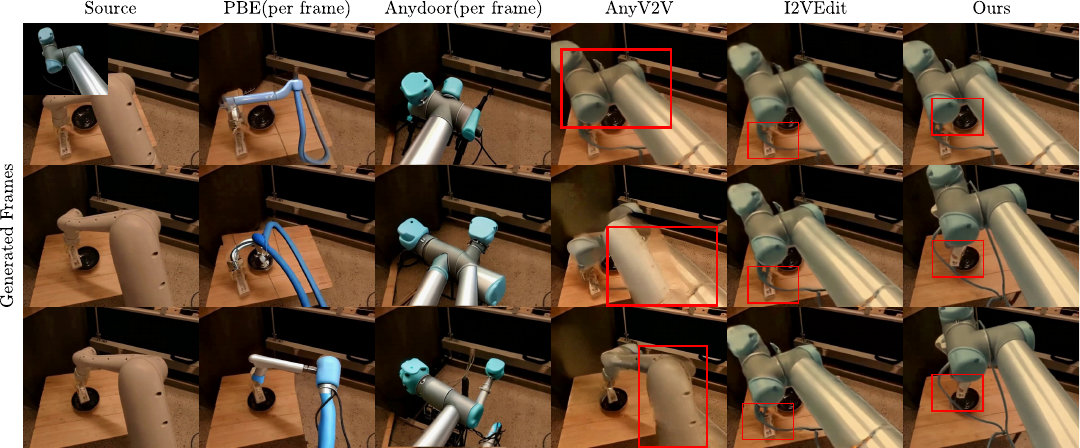}
  \caption{Comparison against different Video Editing Approaches for Robot Swap on the Google Robot → UR5 Robot task.}
  \label{fig:main_a2e}
\end{figure*}

\begin{figure*}[ht!]
  \centering
  \includegraphics[width=1\textwidth]{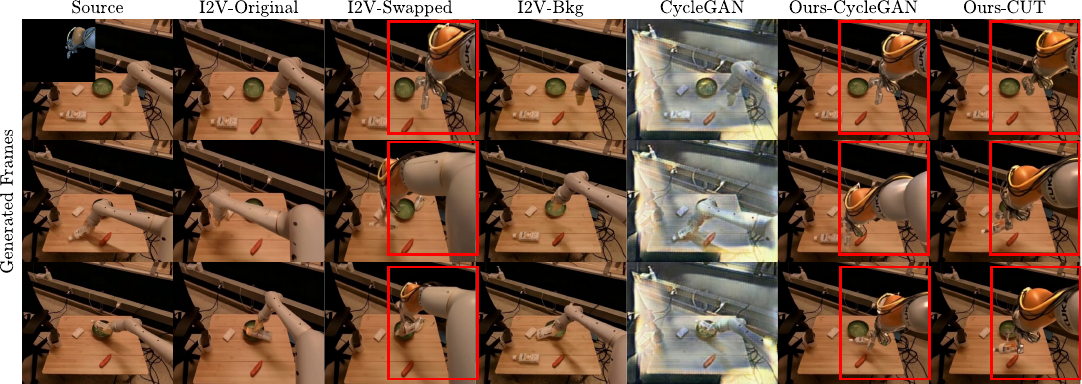}
  \caption{Ablation study against different approaches for Robot Swapping on the Google Robot $\rightarrow$ Kuka task.}
  \label{fig:ablations}
  \vspace{-2em}
\end{figure*}

\noindent\textbf{Qualitative evaluation} For Image Editing, from Fig.~\ref{fig:imgedit_comparison} we can see that our method demonstrates strong replacement capabilities, successfully replacing the appearance of the robotic arm and seamlessly integrating it into the scene by adjusting to the lighting conditions. In contrast, baseline methods such as PBE \cite{yang2023paint} and AnyDoor \cite{chen2024anydoor} fail to achieve a successful replacement. Both inpainting methods generate unrealistic results. In the human evaluation Table \ref{tab:image_comparison}, all participants agreed that our method successfully replaces the robotic arm, whereas other methods were unable to achieve this. 

For video editing, Fig. \ref{fig:main_comp_qual} and \ref{fig:main_a2e} present the edited video frames by RoboSwap compared to the baselines. In \ref{fig:main_comp_qual}, the Google robotic arm is replaced with a Kuka robotic arm, while in \ref{fig:main_a2e}, it is swapped with a UR5 robotic arm. The two image-based baseline methods, which perform frame-by-frame replacement, produce very inconsistent videos and fail to complete the replacement successfully. Among the two tested video-editing baseline methods, we use the best first-frame replacement (generated by our method) as the reference. AnyV2V, a tuning-free approach, fails to maintain the structural consistency of the robotic arm, even though it successfully interacts with the background in the generated video. I2VEdit, which trains a LoRA model\cite{hu2022lora} specifically for the source video, fails to interact with the background and complete the task as objects are not grasped. Additionally, I2VEdit struggles to maintain structural consistency. In contrast, our RoboSwap method successfully replaces the robotic arm while maintaining plausible interactions with objects in the scene. The generated lighting and shadows are also realistic and well-integrated.

\subsection{Ablation Study} In order to showcase the importance of our novel contributions and design choices, we ablate different parts of the pipeline. The baselines based on top of CogVideoX or CycleGAN are considered as ablations and reported in the lower half of Table \ref{tab:main_comparison}.

From Table \ref{tab:main_comparison} and Fig.~\ref{fig:ablations}, we can find that, although GAN-based methods are well-established for unpaired training, they still struggle to perform arm replacement when there are significant scene inconsistencies, highlighting the limitations of pixel-to-pixel GAN approaches. On the other hand, to highlight the importance of our design choice in Stage 2 (conditioning the video diffusion model on the composited foreground from CycleGAN and background video), we compare with I2V-Original, I2V-Bkg and I2V-Swapped. We find that these variants fail to successfully and stably replace the robotic arm. After merely modifying the text prompt, the output video still retains the original robotic arm structure. Even when both the swapped first frame and changed prompt are given, shows only the Google robot can transfer to Kuka successfully shown in Table \ref{tab:main_comparison} and Fig \ref{fig:ablations} shows the I2V-Swapped becomes highly unstable, in most case the resulting video gradually exhibits the characteristics of the original robotic arm. In contrast, our method consistently maintains the structural integrity of the robotic arm while successfully accomplishing the original task. This limitation underscores the shortcomings of powerful video generation models in controllable generation tasks, suggesting significant potential for future research in this area. Note however that the metrics from VBench only assess the quality of the video, not whether the robot arm has been successfully replaced or not. Some ablation experiments can achieve higher results without swapping the arm. In Table \ref{tab:main_comparison}, we add a column to indicate whether the arm has been replaced or not. The qualitative results in Fig \ref{fig:ablations} show the superiority of our method. 
To address this challenge, we successfully integrated the strengths of both models by using the GAN model to generate pseudo video pairs as reference conditions, enabling stable and controlled robotic arm replacement.

\section{CONCLUSIONS}
In this paper, we propose RoboSwap, a novel data-driven approach that combines GANs and diffusion models to address the limitations of current methods in unpaired robotic arm swapping in videos. RoboSwap uses a two-stage design: the first stage translates one robotic arm to another through a GAN model. In the second stage, we train a video diffusion model to inpaint a given robotic arm into the video stream of a given environment. Finally, in the inference stage, we input the videos from stage 1 into the diffusion model to obtain the results.


Both qualitative and quantitative results demonstrate that RoboSwap successfully maintains the motion dynamics of the original robot while producing realistic videos. This work expands the capabilities of generative models in video editing for embodied intelligence, with potential benefits in increasing the scale and diversity of robotic video datasets, reducing manual data collection efforts, and enabling large-scale robotic learning across various tasks, environments, and embodiments. 

Swapping robotic arms in unpaired videos remains a challenge. While RoboSwap represents a promising first step, we identify the following areas for improvement in future works: (1) enhancing robotic arm-object interaction by incorporating metadata such as optical flow, and motion trajectories, (2) generalization across robotic arms by developing a one-to-many model that can generalize across different robotic arms without requiring retraining for every two robotic arms. Future works can focus on addressing these challenges to improve the robustness and scalability of RoboSwap.

{
\bibliographystyle{IEEEtran}
\bibliography{IEEEabrv, references}
}









\end{document}